Article information

## Article title

Synthetic Distracted Driving (SynDD2) dataset for analyzing distracted behaviors and various gaze zones of a driver.

## Authors
Mohammed Shaiqur Rahman [a] *, Jiyang Wang [b], Senem Velipasalar Gursoy [b], David Anastasiu [c], Shuo Wang [d], Anuj Sharma [a]

## Affiliations
[a] Iowa State University, Ames, Iowa
[b] Syracuse University, Syracuse, New York
[c] Santa Clara University, California
[d] NVIDIA Corporation, California

## Corresponding author's email address and Twitter handle
shaiqur@iastate.edu (M.S. Rahman), jwang127@syr.edu (J. Wang), svelipas@syr.edu (S. V. Gursoy), danastasiu@scu.edu (D. Anastasiu), shuow@nvidia.com (S. Wang), anujs@iastate.edu (A. Sharma)

## Abstract
This article presents a synthetic distracted driving (SynDD2 - a continuum of SynDD1 [1]) dataset for machine learning models to detect and analyze drivers' various distracted behavior and different gaze zones. We collected the data in a stationary vehicle using three in-vehicle cameras positioned at locations: on the dashboard, near the rearview mirror, and on the top right-side window corner. The dataset contains two activity types: distracted activities [2][3][4] and gaze zones [5][6][7] for each participant, and each activity type has two sets: without appearance blocks and with appearance blocks such as wearing a hat or sunglasses. The order and duration of each activity for each participant are random. In addition, the dataset contains manual annotations for each activity, having its start and end time annotated. Researchers could use this dataset to evaluate the performance of machine learning algorithms to classify various distracting activities and gaze zones of drivers.

**Specifications table**

| | |
|---|---|
| **Subject** | Data Science |
| **Specific subject area** | Driver behavior analysis, Driver safety |
| **Type of data** | InfraRed Videos, Annotation files |
| **How the data were acquired** | Three in-vehicle cameras acquired data. We requested the participants to sit in the driver's seat and then instructed them to perform driver-distracting activities or gaze at some region for a short time interval. The instructions were played on a portable audio player.<br><br>Instruments: Kingslim D1 dash cam [8] |
| **Data format** | Video files are .MP4 format and annotation files are .csv files |
| **Description of data collection** | We designed a survey using a Qualtrics form and selected the respondents based on the criteria that created a balanced representation by gender, age, and ethnicity. |
| **Data source location** | · Institution: Iowa State University<br>· City/Town/Region: Ames, Iowa<br>· Country: USA<br>· Latitude 42.0267° N, Longitude 93.6465° W |

**Value of the data**
- The data will serve as baseline data for the training, testing, and validation of computer vision-based machine learning models having the primary objective of detecting and classifying driver behaviors and gaze zones.
- The data can be used to benchmark the performance of various machine learning models designed with a similar objective.
- The data can be used by researchers working on analyzing driver behaviors whose objective is detecting and classifying driver activities.
- The data can help researchers design and build a driver-assist system to improve drivers' safety by alerting them during driving.

# 1. Data description

The dataset consists of video files and annotation files. The videos were collected using dashcams with the specification shown in Table 1, and the data acquisition requirements are shown in Table 2. The annotation files(.csv) contain information, as shown in Table 3, and it includes information for all the camera views, as shown in Figure 1.

| Camera model | Kingslim D1 Pro Dual Dash Cam |
|---|---|
| Resolution | 1920 x 1080P |
| Frame rate | 30fps |
| Sensor | 1/ (2.8)" SONY IMX307 industrial grade |
| Aperture | f/1.8 large |
| Lens angle | 170-degree wide angle |
| LED | 4 IR LED's |
| Pixel size | $2.9\mu$ x $2.9\mu$ |

Table 1. Showing Specifications of the video acquisition system

| Requirement | Description |
|---|---|
| Data Recording | Video duration: 300 s multiple videos |
|  | Frequency: 50hz |
|  | Data file format: MP4 |
| Data Storing | SD card reader: Sandisk micro |
| Operating system | Windows 10 (& above), Mac OS Sierra (& above) |
| Communication | USB 2, USB C |

Table 2. Showing Data acquisition requirement

| User ID | Participant identification number |
|---|---|
| Filename | Video file name |
| Camera View | Camera positioned near the dashboard, rearview mirror, or right-side window |
| Activity Type | Gaze or distracted |
| Start Time | Start time (h: mm: ss) of activity (may include up to 10 seconds before the actual activity starts) |
| End Time | End time (h: mm: ss) of activity |
| Label | Various activities performed by the participant; see table 3 and 4 for more details |
| Appearance Block | Participants may be wearing hat or sunglass or none |

Table 3. Showing variables in the data set

For each participant, there are twelve video files since each camera has two activity types (gaze/distracted), and each activity type has two sets (with/without appearance block), as shown in Table 4. The video files are infra-red, and they do not contain any audio data.

| Dashboard | Gaze | Without appearance block |
| --- | --- | --- |
| | | With appearance block |
| | Distracted | Without appearance block |
| | | With appearance block |

Table 4. Showing different videos for one camera view

## 2. Experimental design, materials, and methods

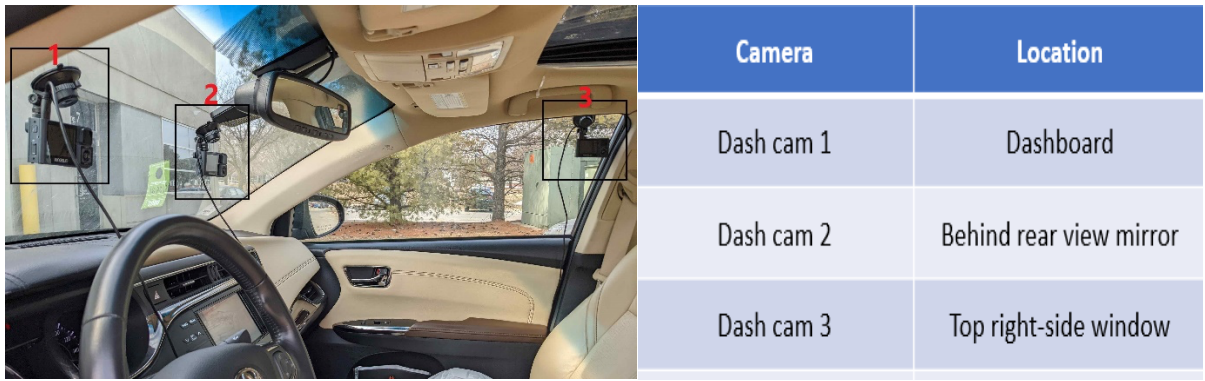

Figure 1. Showing camera positions inside the car

The synthetic data collection process involved three in-vehicle cameras [8] positioned near the dashboard, rearview mirror, and top corner of the right-side window, as shown in Figure 1. We requested the participants to sit inside a stationary vehicle in the driver's seat. Then we gave them instructions (played using a tablet) to gaze at a particular region or perform a distracting activity continuously for a short time. The duration and order of activities were random for each participant. The dataset, thus generated, we call Synthetic Distracted Driving (SynDD2).

The specification of the videos in the dataset is shown in Table 6.

| Format | .MP4 |
| --- | --- |
| Video codec | H.264/AVC |
| Framerate | 30FPS |
| Video bitrate | 11.88Mbit/s |
| Resolution | 1920 × 1080 |
| Aspect ratio | 16:9 |

Table 6. Showing specifications of videos

## 2.1 Gaze zone

Figure 2 shows the eleven gaze zones in the car, and Table 7 lists all the gaze zones.

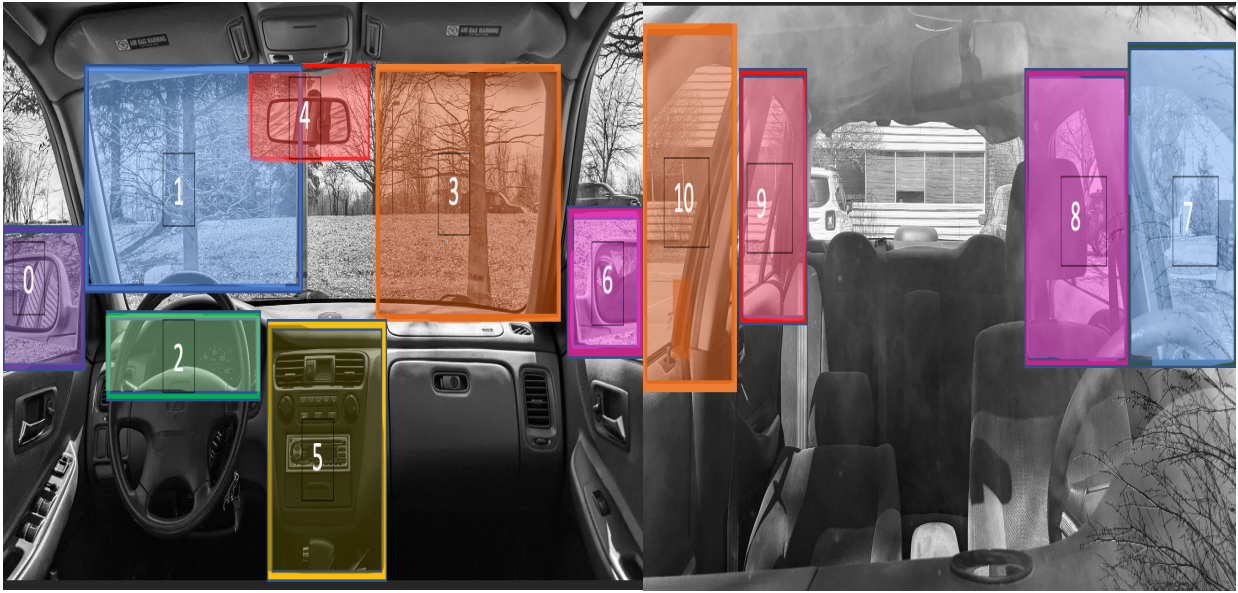

Figure 2. Showing the gaze zones/regions

| S. No | Gaze zones/regions |
|---|---|
| 0 | left-rear mirror |
| 1 | forward window |
| 2 | speedometer |
| 3 | right-frontal window |
| 4 | rear mirror |
| 5 | control panel and shift |
| 6 | right-rear mirror |
| 7 | left-side window |
| 8 | left blind spot |
| 9 | right-side blind spot |
| 10 | right-side window |

Table 7. Showing gaze zones

## 2.2 Distracted behavior

Each participant continuously performed sixteen distracted driver behavior, as shown in figure 3, for a small-time interval. We have listed the sixteen activities in Table 8.

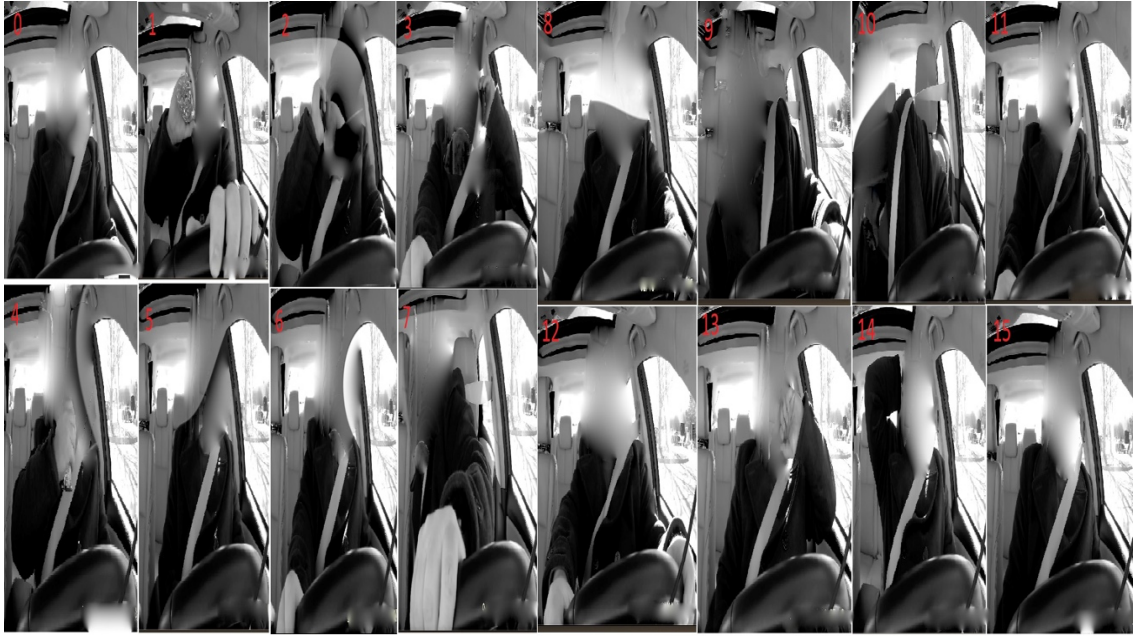

Figure 3. Showing distracted behaviors

| S. No | Distracted driver behavior |
|---|---|
| 0 | Normal Forward Driving |
| 1 | Drinking |
| 2 | Phone Call(right) |
| 3 | Phone Call(left) |
| 4 | Eating |
| 5 | Text (Right) |
| 6 | Text (Left) |
| 7 | Reaching behind |
| 8 | Adjust control panel |
| 9 | Pick up from floor (Driver) |
| 10 | Pick up from floor (Passenger) |
| 11 | Talk to passenger at the right |
| 12 | Talk to passenger at backseat |
| 13 | Yawning |
| 14 | Hand on head |
| 15 | Singing or dancing with music |

Table 8. Showing distracted behavior

## 2.3 Method

We requested the participants to follow the instructions, which were played on a portable audio player, and after the completion of one set of activities, we requested them to repeat the set by wearing a hat or sunglasses. One set of gaze activities took approximately 5-6 minutes, while the distracted driving activities took around 10 minutes. The whole set of activities took about one hour to finish.
The sequence and duration of each activity were randomized for each participant to introduce complexity in the data for analysis.

## 2.4 Instructions for activities

We created an instruction video in the English language for both activity types for each participant. We used gTTS [9] (Google Text-to-Speech) to convert the instructions into text-to-speech form. The video showed the region to gaze for the gaze activity type, and for the distracted activity type, it displayed activity names in the English language, like drinking, eating, etc. The instructions started by explaining the kind of activity the participant would perform. Then the instruction video played a small beep sound for the participant to begin acting. They continued to gaze until they heard a long beep sound for the gaze activity type, and for the distracted activity type, we requested them to act naturally: to stop performing whenever they wanted or when they heard a long beep.
    We added the beep sounds to synchronize the videos from different camera views and help annotate the activities manually.

## 2.5 Data pre-processing

By default, each camera would split the video files after a fixed time interval. As a result, each participant's raw data had multiple video files from a single camera. Hence, we combined each participant's video files into a single file using python and FFmpeg [10].

"ffmpeg -f concat -safe 0 -i video-input-list.txt -c copy 'output'"

where:
video-input-list.txt- video files from a camera view in the increasing order of time
output- the name of the output file

We sorted the video files and added the file names in the video-input-list.txt file. Then using FFmpeg, we concatenated the videos listed in the text file giving us a single file output.

After that, we divided the output video into multiple video files based on the activity types: gaze, gaze with appearance block, distracted, and distracted with appearance block for each camera view.

"ffmpeg -ss {start} -t {dur} -i {p} -c copy {out}"

where:
{start} - represents the start time of the activity type (gaze/distracted),
{dur} - represents the length of that activity type,

{p} - represents the path of the input file, and
{out} - represents the output file name.

Finally, we synchronized the videos from the three camera views based on the beep sound played in the instruction.

### 2.6 Data annotation

We annotated each video from each camera for each participant manually. The annotation file includes each activity's start and end times—more information is in Table 1.

### Ethics statements

We got approval from IRB and then started the data collection process. We confirm that each participant signed the IRB-approved informed consent form before collecting data. The consent form clearly states that the data (showing the face) will be used in data challenges and competition and will be released for worldwide use as the data set. For Figure 2 and Figure 3, we confirm that we got the IRB-approved consent from the participants.

### CRediT author statement

Mohammed Shaiqur Rahman: Writing- Original draft preparation, Data curation, Software, Methodology, Investigation, Design. Jiyang Wang: Formal analysis, Study Inception, and Design. Anuj Sharma: Study Inception and Design. Senem Velipasalar Gursoy: Study Inception and Design. David Anastasiu: Study Inception and Design. Shuo Wang: Study Inception and Design.

### Acknowledgments

Our research results are based upon work supported by the U.S. Department of Transportation Exploratory Advanced Research Program under Award No. 693JJ31950022. Any opinions, findings, conclusions, or recommendations expressed in this material are those of the author(s) and do not necessarily reflect the views of the U.S. Department of Transportation.

### Declaration of interests

The authors declare that they have no known competing financial interests or personal relationships that could have appeared to influence the work reported in this paper.

**References**

[1] M. S. Rahman *et al.*, "Synthetic distracted driving (SynDD1) dataset for analyzing distracted behaviors and various gaze zones of a driver," *Data Brief*, vol. 46, p. 108793, Feb. 2023, doi: 10.1016/j.dib.2022.108793.
[2] State Farm Distracted Driver Detection Kaggle Challenge (link)
[3] Martin, Manuel, et al. "Drive&act: A multi-modal dataset for fine-grained driver behavior recognition in autonomous vehicles." Proceedings of the IEEE/CVF International Conference on Computer Vision. 2019.
[4] Seshadri, Keshav, et al. "Driver cell phone usage detection on strategic highway research program (shrp2) face view videos." Proceedings of the IEEE Conference on Computer Vision and Pattern Recognition Workshops. 2015. (link)
[5] Rangesh, Akshay, Bowen Zhang, and Mohan M. Trivedi. "Driver gaze estimation in the real world: Overcoming the eyeglass challenge." 2020 IEEE Intelligent Vehicles Symposium (IV). IEEE, 2020.
[6] Vora, Sourabh, Akshay Rangesh, and Mohan M. Trivedi. "On generalizing driver gaze zone estimation using convolutional neural networks." 2017 IEEE Intelligent Vehicles Symposium (IV). IEEE, 2017.
[7] Vora, Sourabh, Akshay Rangesh, and Mohan Manubhai Trivedi. "Driver gaze zone estimation using convolutional neural networks: A general framework and ablative analysis." IEEE Transactions on Intelligent Vehicles 3.3 (2018): 254-265.
[8] https://kingslim.net/products/kingslim-d1-pro-dual-dash-cam
[9] https://gtts.readthedocs.io/en/latest/
[10] https://www.ffmpeg.org/ffmpeg.html